\pdfoutput=1
\documentclass[11pt]{article}

\usepackage[preprint]{acl}

\usepackage{multicol}
\usepackage{multirow}
\usepackage{array}
\usepackage{booktabs}
\usepackage{tabularx}
\usepackage{threeparttable}
\renewcommand{\arraystretch}{1.08}
 \usepackage{amssymb}
\usepackage{times}
\usepackage{latexsym}
\usepackage{amsmath}
\usepackage[utf8]{inputenc}
\usepackage[T1]{fontenc}
\usepackage{microtype}
\usepackage{inconsolata}
\usepackage{graphicx}
\usepackage{url}

\usepackage{tikz}
\usetikzlibrary{shapes.geometric, arrows.meta, positioning, calc, fit}
\usepackage{xcolor}

\definecolor{categoryblue}{RGB}{220, 235, 252}
\definecolor{categorygreen}{RGB}{220, 245, 230}
\definecolor{categoryorange}{RGB}{255, 235, 215}
\definecolor{categorypurple}{RGB}{240, 225, 250}
\definecolor{headerdark}{RGB}{40, 60, 90}
\definecolor{borderblue}{RGB}{100, 140, 200}

\title{Medical Reasoning in the Era of LLMs: A Comprehensive Review of Enhancement Techniques and Applications}

\author{
\textbf{Zizhan Ma}$^{1}$ \quad \textbf{Wenxuan Wang}$^{2*}$ \quad \textbf{Meidan Ding}$^{3}$ \quad \textbf{Shiyi Zheng}$^{3}$ \quad \textbf{Shengyuan Liu}$^{1}$ \\ 
\textbf{Jie Liu}$^{4}$ \quad \textbf{Jiaming Ji}$^{5}$  \quad \textbf{Linlin Shen}$^{3}$ \quad \textbf{Yixuan Yuan}$^{1}$ \quad \textbf{Wenting Chen}$^{6}$\thanks{Wenxuan Wang and Wenting Chen are corresponding authors.}\\
$^1$ The Chinese University of Hong Kong \quad \quad $^2$ Renmin University of China \quad \quad $^3$ Shenzhen University\\
 $^4$ City University of Hong Kong \quad \quad $^5$ Peking University   \quad \quad $^6$ Stanford University\\
{\fontsize{9.5pt}{11pt}\selectfont \texttt{zizhan.ma@link.cuhk.edu.hk} \quad \texttt{wangwenxuan@ruc.edu.cn} \quad \texttt{wentichen7-c@my.cityu.edu.hk}}
}

\begin{document}
\maketitle

\begin{abstract}
Large Language Models (LLMs) demonstrate strong capabilities in medical text processing, yet their capacity for verifiable, multi-step medical deduction remains limited. This limitation has motivated a shift from direct answer generation toward specialized architectures for medical reasoning. We present a dual taxonomy categorizing enhancement methodologies into training-time strategies (supervised fine-tuning, reinforcement learning) and test-time mechanisms (structured prompting, reasoning selection, knowledge integration, multi-agent orchestration), aligned with clinical cognitive execution stages. We analyze these techniques across text, vision, and code modalities within core clinical workflows, including diagnosis, treatment planning, and drug discovery. Furthermore, we trace the transition from static answer accuracy to dynamic process-level evaluation and multimodal interpretability. Synthesizing research from 2022 to 2026, this review identifies persistent challenges—such as plausible hallucinations, conflicts between parametric knowledge and dynamic evidence, multimodal alignment gaps, and efficiency-performance trade-offs—and outlines actionable directions for robust, interpretable medical AI.
\end{abstract}

\section{Introduction}

Specialized foundation models process biomedical knowledge effectively \cite{Medpalm2, wu2023pmcllama, luo2022biogpt, zhou2023survey}, yet struggle when executing multi-step inference for complex clinical decisions \cite{sim2025critique, kim2025limitations}. Medical diagnosis requires causal reasoning to integrate fragmented symptoms, patient history, and diagnostic test results into a coherent clinical formulation \cite{richens2020improving, xue2024ai, mattingly1991clinical}. Diagnostic inaccuracies remain a major driver of clinical malpractice claims, highlighting the risks of flawed medical deduction \cite{Rates}.

Structured prompting techniques extract intermediate reasoning steps to improve model output transparency \cite{wei2023chainofthoughtpromptingelicitsreasoning, nachane2024few}. Building on these techniques, recent research develops architectures that simulate diagnostic workflows, verify intermediate findings, and adapt to ambiguous patient presentations \cite{savage2024diagnostic, kim2024mdagents}. These systems support clinical decision-making and medical education by providing feedback that improves diagnostic accuracy \cite{llmimprovedecision, llmsupport}. Recent developments embed verifiable logic into model outputs through reinforcement learning and knowledge graph constraints, transitioning LLMs from passive retrieval engines to structured deductive agents \cite{pan2025survey, MedReason}.

We explicitly distinguish medical reasoning from standard clinical text mining or knowledge retrieval based on the requirement of multi-step deductive path generation. Tasks such as automated treatment planning and computational drug discovery require more than static information retrieval; they demand multi-turn logic trees, counterfactual hypothesis testing, and interactive state tracking within simulated clinical sandboxes \cite{xu2025medagentgym, liu2024drugagent}.

Our systematic screening protocol queried PubMed, Scopus, and arXiv databases using the search string: \texttt{(("clinical reasoning" OR "medical logic" OR "deductive enhancement") AND ("large language models" OR "LLM" OR "agents"))}. From records published between January 2022 and March 2026, studies were included if they proposed post-training data/reward adjustments or test-time scaling strategies that explicitly optimize multi-step logical traces. General medical text mining literature unrelated to multi-step reasoning or decision-making workflows was excluded, yielding a curated body of core studies spanning enhancement techniques, clinical applications, and evaluation frameworks.

This review extends prior capability and evaluation surveys \cite{zhou2023survey, shool2025systematic} by focusing directly on the algorithmic mechanics of reasoning enhancement and its associated structural limitations \cite{kim2024epistemology, soffer2025pitfalls}. To clarify our scope relative to existing literature \cite{zhou2023survey, shool2025systematic, berger2025reasoning, mansoor2026reasoning, peng2025aligning, lin2026survey, ren2026mrbench}, we provide a structured Survey Gap Table in Appendix Table~\ref{tab:survey_comparison}.

The review is structured as follows: Section~\ref{sec:background_framework} introduces modal requirements, clinical cognitive stages, and our dual taxonomy; Section~\ref{sec:architectures} analyzes training-time and test-time enhancement methods; Section~\ref{sec:applications} examines clinical applications; Section~\ref{sec:evaluation} surveys evaluation frameworks; Section~\ref{sec:discussion} analyzes remaining barriers to clinical adoption.

\section{Background and Conceptual Framework}
\label{sec:background_framework}

\subsection{From General Foundation Models to Specialized Medical Reasoning}
Medical reasoning synthesizes clinical evidence to formulate accurate diagnoses and actionable management plans. Although general LLMs excel at processing clinical text, standard probabilistic architectures are not inherently optimized for the structured, multi-step inference required in high-stakes clinical settings. Operating as next-token predictors, standard LLMs often struggle with formal causal operations, such as differentiating correlation from causation or quantifying clinical uncertainty \cite{wei2023chainofthoughtpromptingelicitsreasoning}. Chain-of-Thought (CoT) prompting demonstrates that generating explicit intermediate steps elicits latent reasoning capabilities \cite{wei2023chainofthoughtpromptingelicitsreasoning, nachane2024few}. Modern specialized reasoners combine supervised fine-tuning on explicit deductive traces with reinforcement learning to systematically reinforce logically sound medical reasoning \cite{openai2024openaio1card, pan2025survey, sonoda2025structured}.

\subsection{Modalities in Medical Reasoning}
\label{sec:modalities}

The underlying data modality dictates specific reasoning requirements and constraints.

\subsubsection{Reasoning over Text}
Clinical text often lacks formalized logic, making coherent inference challenging. Training strategies make internal steps explicit, producing rationales aligned with clinical patterns \cite{sft/EMULATION} or structured knowledge graphs \cite{MedReason}. Reasoning trace validity can be enforced via first-order logic \cite{zafar2025medex} or reinforcement learning that rewards factual correctness and penalizes ungrounded generation \cite{Med-RLVR, liu2025beyond}. Inference-time techniques deepen reasoning through test-time compute scaling \cite{huang2025m1, FineMedLM-o1, shi2024medadapter} or broaden perspective through multi-agent consensus debate \cite{hong2024argmed, tang2024medagents, zhu2025ask}.

\subsubsection{Reasoning over Medical Images}
Medical image reasoning bridges high-dimensional pixel patterns with abstract diagnostic categories. Visual grounding is essential for logical consistency and clinician trust. Early visual-language integration applied reinforcement learning to align visual features directly with diagnostic labels \cite{pan2025medvlm, Med-R1, GMAI-VL-R1, zhang2025medtvt}. Two-stage approaches like MC-CoT \cite{wei2024mc} and Elicit and Enhance \cite{mu2025elicit} utilize language models to plan visual analysis steps executed by Vision-Language Models (VLMs). Advanced methods enforce spatial grounding by incorporating anatomical location constraints \cite{trinh2025prsmed} or anatomical ontologies to formalize image interpretation \cite{AOR}.

\subsubsection{Reasoning over Biomedical Code}
Code-based reasoning supports programmatic, verifiable clinical workflows. Datasets like BioCoder \cite{tang2023biocoder} establish that public biomedical code can support training for specialized code generation. Execution sandboxes such as MedAgentGym \cite{xu2025medagentgym} provide standardized environments for agent training. Benchmarks like BioDSBench \cite{wang2024can} integrate validated code agents into clinical data science pipelines, establishing a clear separation between agent training environments and end-user clinical software.

\begin{figure*}[t!]
    \centering
    \includegraphics[width=\textwidth]{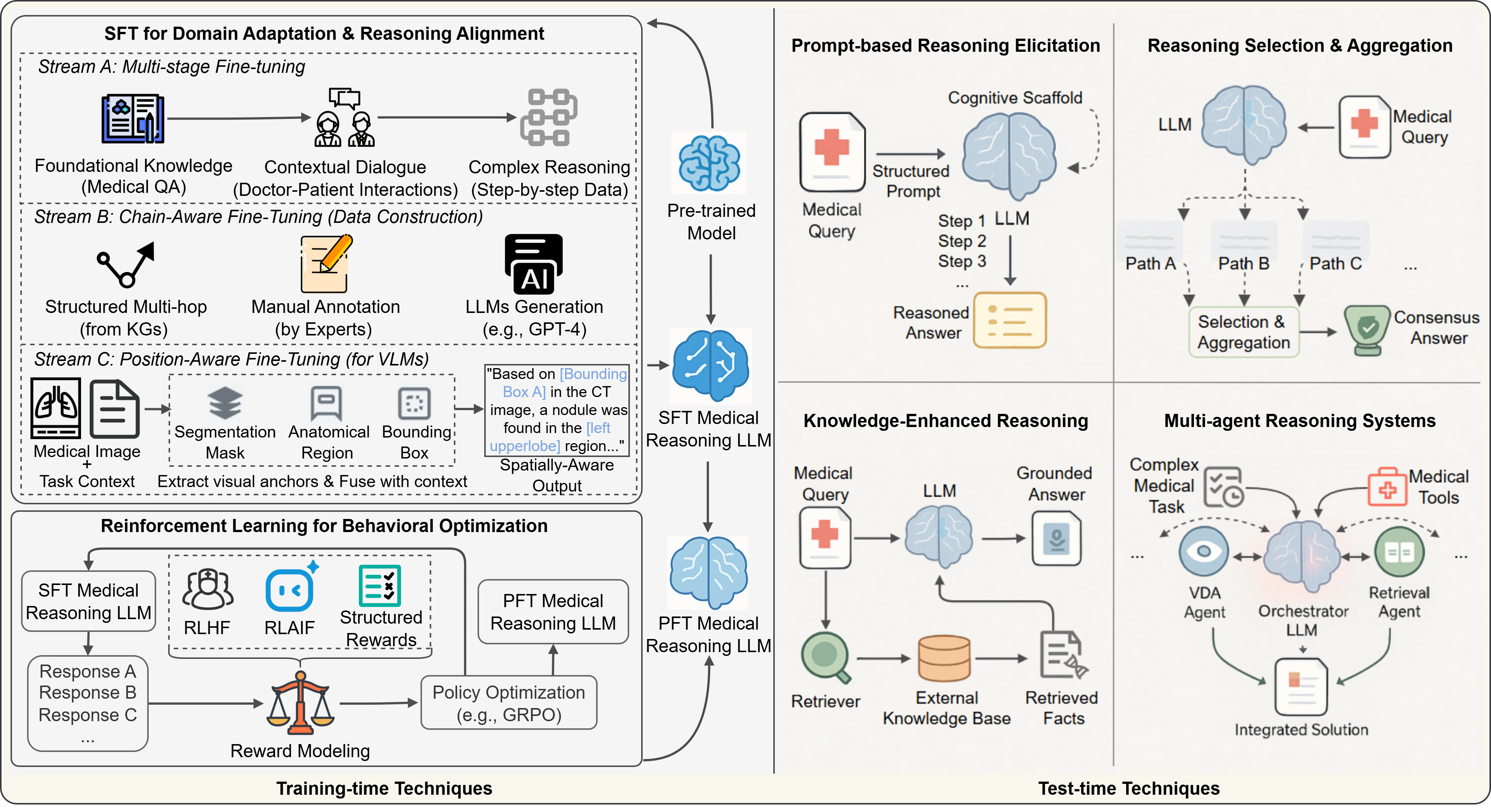}
    \caption{
        An overarching framework of techniques to enhance medical reasoning in LLMs. Training-time Techniques instill models with deductive capabilities by adjusting internal weights via SFT and RL. Test-time Techniques enhance deduction at inference, encompassing Prompt-based Elicitation, Reasoning Selection and Aggregation, Knowledge-Enhanced Reasoning, and Multi-agent Systems.
    }
    \label{fig:overall_framework}
\end{figure*}

\subsection{A Clinical Cognitive Perspective in Medical Reasoning}
\label{sec:conceptual_framework}

Medical reasoning directly maps onto the cognitive stages of clinical practice: (1) Hypothesis Generation \& Differential Diagnosis, where clinicians evaluate patient presentations to form initial differential diagnostic lists \cite{kwon2023large, tang2024medagents}; (2) Evidence Gathering \& Hypothesis Refinement, an iterative process where targeted diagnostic testing and patient queries systematically discriminate among competing diagnoses \cite{he2024bp4er, zafar2025medex}; and (3) Synthesis \& Verification, where accumulated findings are integrated into a coherent diagnostic narrative confirming the final assessment and management strategy \cite{AOR, Reasonmed}. A complete breakdown mapping these cognitive stages to specific LLM mechanisms is provided in Appendix Table~\ref{tab:cognitive_stages}.

To categorize computational interventions supporting these cognitive stages, we establish a dual taxonomy that divides enhancement methodologies into two primary phases: Training-time Techniques (modifying internal weights via SFT and RL to embed clinical logic directly into parameters) and Test-time Techniques (guiding inference outputs via prompting, candidate selection, knowledge retrieval, and multi-agent systems without weight updates), as shown in Figure~\ref{fig:overall_framework}.

By uniting modality requirements, training/test-time execution phases, and clinical cognitive stages, this framework connects machine learning optimization directly with real-world clinical practice.

\section{A Taxonomy of Medical Reasoning Enhancement Techniques} \label{sec:architectures}

Methodologies to enhance medical reasoning divide into training-time and test-time execution phases (Figure~\ref{fig:overall_framework}).

\subsection{Training-time Techniques: Building the Foundation}

Training-time methods embed clinical domain logic directly into model weights, though they depend heavily on high-quality annotated data.

\subsubsection{Supervised Fine-tuning (SFT)}
Supervised fine-tuning transitions models from surface pattern matching to procedural medical reasoning. Curriculum-based multi-stage fine-tuning decomposes reasoning into structured learning steps. Task-abstraction staging builds capabilities from factual retrieval to complex multi-step deduction, as demonstrated by FineMedLM-o1 \cite{FineMedLM-o1}. Modality-integration staging manages multimodal tasks, advancing sequentially from perceptual feature extraction to clinical interpretation \cite{AOR}.

Chain-aware fine-tuning depends on high-quality step-by-step rationales. While human expert annotations provide accurate supervision \cite{Medpalm2}, automated rationale generation offers scalable data expansion. Frameworks like HuatuoGPT-o1 \cite{HuatuoGPT-o1} utilize teacher models with self-correction loops, though this risks transferring teacher biases. Verifiable methods impose explicit structural constraints; for example, MedReason \cite{MedReason} restricts rationale paths to verified knowledge graph traversals for auditability. Other approaches clean training datasets using quality filtering \cite{sft/BioMed-R1} or style alignment \cite{sft/EMULATION}.

Position-Aware Fine-tuning (PFT) links textual diagnostic assertions directly to visual image regions, establishing spatial grounding. Coarse-grained anchoring uses bounding box coordinates during intermediate reasoning \cite{MedGroundR1}. Fine-grained anchoring uses pixel-level segmentation masks for high-precision localization \cite{trinh2025prsmed}. Semantic anchoring connects image regions directly to standardized medical vocabularies \cite{AOR}.

SFT models are sensitive to rationale quality. Fine-tuning on unverified CoT data generated by general-purpose LLMs \cite{sft/BioMed-R1, FineMedLM-o1} can lead to superficial reasoning traces that appear fluent but contain logical flaws. Consequently, automated rationale verification and data-centric filtering pipelines are critical to remove faulty reasoning paths prior to fine-tuning.

\subsubsection{Reinforcement Learning (RL)}
Reinforcement learning aligns post-training outputs with clinical requirements. Med-PaLM 2 \cite{Medpalm2} applied physician ranking preferences via RLHF to reinforce clinical safety and accuracy. To improve scalability, RLAIF uses advanced models (e.g., GPT-4o in HuatuoGPT-o1 \cite{HuatuoGPT-o1}) to generate scalar rewards. Structured reward functions optimized via Group Relative Policy Optimization (GRPO) \cite{GRPO} enable objective, reproducible optimization across composite metrics:
\begin{equation}
R_{total} = \alpha \cdot R_{accuracy} + \beta \cdot R_{logic} + \gamma \cdot R_{grounding}
\end{equation}
where $R_{accuracy} \in \{0, 1\}$ evaluates terminal answer correctness against ground-truth clinical targets; $R_{logic} \in [0, 1]$ assesses the structural validity and causal coherence of intermediate reasoning steps via rule-based format validators (e.g., checking tag formatting like \texttt{<think>...</texttt>} or logic step syntax); and $R_{grounding} \in [0, 1]$ measures spatial alignment in multimodal settings, calculated via Intersection over Union ($\text{IoU}(B_{pred}, B_{gt})$) or Dice scores comparing predicted visual regions $B_{pred}$ against expert anatomical masks $B_{gt}$.

Furthermore, GRPO optimizes policy models by comparing candidate trajectories within a sampled group against their relative baseline, eliminating the need for a separate critic model and significantly lowering GPU memory overhead during medical reasoning alignment. This group-relative advantage estimation ensures stable gradient updates even when rewarding dense, multi-step rationales. While objective rewards allow complex reasoning patterns to emerge without explicit rationale distillation \cite{liu2025beyond, sft/BioMed-R1}, RL remains sensitive to reward design. Incomplete or poorly calibrated reward metrics can cause reward hacking, where models modify output formatting to maximize rule compliance without improving actual medical logic correctness.

\subsubsection{Challenges in Training-time Optimization}
SFT for medical deduction is fundamentally constrained by expert annotation costs and risks stylistic emulation \cite{sft/EMULATION}, where models replicate linguistic templates without learning causal relationships \cite{chu2024professionalagentsevolving}. Meanwhile, RL optimization faces reward brittleness, as simple exact-match or format-based rewards cannot capture complex clinical safety demands.

\subsection{Test-time Techniques: Achieving Agility and Verifiability}
\label{sec:test_time}

Test-time techniques guide inference trajectories dynamically without updating underlying model parameters.

\subsubsection{Prompt-based Reasoning Elicitation}
Structured prompting elicits latent reasoning paths, progressing from standard CoT \cite{nachane2024few} to clinical variations including Clinical CoT \cite{kwon2023large}, Diagnostic Reasoning CoT \cite{wu2024large}, and Chain of Diagnosis \cite{chen2024cod}. Least-to-most prompting decomposes complex cases into manageable sub-problems \cite{he2024bp4er}. Other methods apply iterative verification steps \cite{vladika2025step}, coordinate cross-modal visual analysis \cite{wei2024mc, zhou2025pathcot}, or execute bidirectional inference to build transparent differential diagnoses \cite{zhou2025dualinf}. Structured clinical prompts enforce systematic workflows from history review to diagnostic formulation, improving rationale interpretability \cite{structured_clinical_2026}. Similarly, evidence-driven prompting pairs context retrieval with explicit evidence scoring to suppress hallucinations \cite{evidence_driven_cot_2026}.

\subsubsection{Reasoning Selection and Aggregation}
\label{sec:inference_refinement}
Evaluating multiple reasoning paths reduces generation variance. In sampling-based inference scaling, self-consistency \cite{wang2023self} and ensemble reasoning \cite{lucas2024reasoning} generate candidate trajectories under higher decoding temperatures and apply majority voting to identify consensus answers. Advanced test-time compute scaling applies Tree-of-Thoughts search \cite{yao2023treethoughtsdeliberateproblem} and search-based inference scaling \cite{huang2025o1replicationjourney, huang2025m1, faray2025does, guo2025deepseek} to explore differential diagnostic branches and self-correct logical errors. Process-level Reward Models (PRMs) \cite{medprm2025} assist search algorithms by scoring intermediate logic steps.

To reduce latency and compute costs during search in acute clinical settings, lightweight post-hoc ranker adapters provide an efficient alternative. Frameworks like MedAdapter \cite{shi2024medadapter} train small adapter modules to score candidate outputs from larger base models based on clinical plausibility. Complementarily, meta-cognitive regulation (e.g., MedCog \cite{medcog2026}) dynamically scales search sampling density based on case difficulty, balancing diagnostic accuracy with operational efficiency.

\subsubsection{Knowledge-Enhanced Reasoning}
\paragraph{Unstructured Knowledge Contextualization} Standard Retrieval-Augmented Generation (RAG) often fails when addressing multi-step clinical queries. Iterative retrieval systems dynamically fetch reference data at specific intermediate steps \cite{yu2023chain}. Frameworks such as SelfRewardRAG \cite{selfrewardrag2024} integrate internal verification loops to score retrieved passages for relevance, filtering out unhelpful context that could trigger hallucinations. Privacy-aware retrieval architectures allow clinicians to control personal health record inclusion during inference \cite{medai_plus_2026}.

\paragraph{Structured Knowledge Guidance} Structured Knowledge Graphs (KGs) enforce physiological constraints during path generation. In-Context Padding \cite{wu2024guiding} inserts structured UMLS triplets into the context window. MedReason \cite{MedReason} validates intermediate steps directly against KG nodes during generation. Hybrid architectures \cite{cabello2024meg} and MedEx \cite{zafar2025medex} combine First-Order Logic (FOL) or clinical guidelines \cite{staniek2025training} to verify consistency between retrieved evidence and diagnostic claims.

\paragraph{Conflict Resolution in Knowledge Integration} Integrating multiple knowledge sources requires resolving contradictions among static clinical guidelines, newly published literature, and patient EHR data. We formalize an asymmetric evidence hierarchy: static medical ontologies act as primary safety boundaries that override retrieved text unless a verified temporal update is confirmed from high-credibility sources \cite{zheng2025kcr}. Formally, an NLI classifier calculates the contradiction probability $P(\text{contradiction} \mid \mathcal{K}_{static}, \mathcal{D}_{dynamic})$ between static knowledge $\mathcal{K}_{static}$ and retrieved text $\mathcal{D}_{dynamic}$. When this contradiction probability exceeds a pre-defined safety threshold $\tau$, the system generates explicit uncertainty markers (\texttt{<uncertainty>}) and executes a safe deferral protocol to transfer decision-making to human expert clinicians \cite{strong2024humanaicollaborationhealthcareguided}.

\subsubsection{Multi-agent Reasoning Systems}
\paragraph{Collaborative Diagnostic Deliberation Agents} ArgMed-Agents \cite{hong2024argmed} structures hypothesis generation, counter-arguments, and evidence updates through formal argumentation schemes. Frameworks like AMANDA \cite{amanda2025} and MultiMedRes \cite{inquire_interact_integrate_2024} use specialized Explorer and Learner components to query data and refine intermediate conclusions for medical visual question answering.

\paragraph{Clinical Workflow Execution Agents} Workflow agents execute specialized operational tasks. MedRAX \cite{fallahpour2025medrax} and Medical AI Consensus \cite{radiologyMA2025} handle radiological workflows across image modality identification, segmentation, and draft report generation. MedChain \cite{liu2026medchain} frames clinical encounters as interactive sequential decision-making tasks to evaluate workflow agents. ClinicalAgent \cite{yue2024clinicalagentclinicaltrialmultiagent} manages clinical trial matching using dedicated patient criteria components, while MedAgentGym \cite{xu2025medagentgym} offers a sandbox environment for code-based clinical deduction.

\paragraph{Diagnostic Quality Evaluation Agents} Agent architectures integrated into Medical AI Consensus, MedAgentGym, AMANDA, and MultiMedRes deploy judge components to score reports, code solutions, or diagnostic decisions based on clinical accuracy, completeness, and guideline compliance.

Multi-agent frameworks risk communication overhead, operational deadlocks, and infinite loops. Assigning distinct proposer and critic roles paired with strict termination conditions (e.g., MedAgents \cite{tang2024medagents}, ArgMed-Agents \cite{hong2024argmed}) ensures agent deliberations reliably converge.

To compare these paradigms, we provide a comparative Methodological Preference Matrix detailing assumptions, failure modes, deployment constraints, and preferred clinical applications in Appendix Table~\ref{tab:methodological_matrix}.

\section{Applications in Clinical Workflow} \label{sec:applications}

\paragraph{History Taking and Patient Representation} Reasoning models process unstructured Electronic Health Records (EHRs) into structured patient representations. EHRAgent \cite{shi2024ehragentcodeempowerslarge} frames tabular EHR reasoning as tool-use planning. Med-Copilot \cite{xu2025medagentgym} extracts clinical information using code-based retrieval steps. AIPatient \cite{yu2024aipatientsimulatingpatientsehrs} combines EHR data with knowledge graphs to simulate realistic patient encounters for clinical education.

\paragraph{Diagnostic Reasoning and Differential Diagnosis} Frameworks including HuatuoGPT-o1 \cite{HuatuoGPT-o1}, MedReason \cite{MedReason}, ReasonMed \cite{Reasonmed}, and Dual-Inf \cite{zhou2025dualinf} generate step-by-step reasoning traces with embedded verification mechanisms, improving diagnostic accuracy \cite{wang2025explainable}. FineMedLM-o1 utilizes trajectory ranking to reduce diagnostic error rates \cite{FineMedLM-o1}. PRS-Med \cite{trinh2025prsmed} improves pathological image analysis, while GRPO-based reinforcement learning \cite{GRPO} enhances reasoning quality across multimodal applications \cite{GMAI-VL-R1, pan2025medvlm, Med-R1, Med-RLVR}. MedGround-R1 \cite{MedGroundR1} incorporates radiological annotations to improve vision-language grounding.

\paragraph{Personalized Treatment Planning} Reasoning models analyze complex, high-dimensional treatment parameters \cite{rao2024proactive, MedWorldModel}. GPT-RadPlan \cite{GPT-RadPlan} assists radiation oncology planning by generating actionable treatment options without domain-specific re-training. MedRBench \cite{MedRBench} benchmarks model performance on the factual accuracy, completeness, and logical consistency of generated treatment rationales.

\paragraph{Prognosis Prediction and Risk Stratification} Identifying key clinical markers improves prognosis assessment and risk stratification accuracy \cite{riskpredict_llm, kim2024llm}. CardioCoT \cite{CardioCoT} introduces a hierarchical reasoning framework for survival analysis, improving interpretability and prediction accuracy for major adverse cardiovascular events in myocardial infarction patients.

\paragraph{Accelerating Drug Discovery and Development} Reasoners assist target identification and molecular optimization pipelines. DrugAgent \cite{liu2024drugagent} frames drug-target interaction prediction as a sequential decision process, demonstrating strong ROC-AUC performance. ChemCrow \cite{ChemCrow} and AgentMol \cite{karabowicz2025agentmol} operate as autonomous agents that integrate computational chemistry tools to plan chemical synthesis pathways and propose candidate molecular structures.

\section{Evaluation \& Benchmarking} \label{sec:evaluation}

\paragraph{Answer Accuracy Assessment} Early benchmarks \cite{jin2020diseasedoespatienthave, pmlr-v174-pal22a, jin2019pubmedqa, hendryckstest2021} focused on multiple-choice accuracy from medical board examinations. Recent evaluations emphasize realistic clinical workflows and distribution shifts \cite{njpdm_eval, chen2026beyond}. MR-Bench \cite{ren2026mrbench} uses real-world hospital data, revealing performance gaps between exam benchmarks and practical clinical accuracy under real-world patient shifts. Specialized datasets evaluate performance in emergency scenarios (ER-Reason \cite{mehandru2026erreason}) and test cognitive flexibility against heuristic biases (mARC \cite{shidara2026marc}). MedXpertQA \cite{medxpertqa} incorporates specialty board questions reviewed by medical experts. MedAgentsBench \cite{medagentsbench} focuses on diagnostic formulation and management planning. Comparative studies evaluating frontier LLMs against physicians on emergency cases show that while LLMs achieve comparable overall accuracy, they frequently miss subtle secondary diagnoses during early differential formulation \cite{medai_plus_2026, usmle_scirep_2026, primellm_2026}.

\paragraph{Reasoning Quality Assessment} Evaluating step-by-step rationale integrity is essential in clinical AI evaluation \cite{nachane2024few, R2MED, zhu2025diagnosisarena}. MedR-Bench \cite{MedRBench} introduces an evaluator that scores rationales across accuracy, completeness, and conciseness. RadRScore \cite{ChestX-Reasoner} measures the factual correctness of intermediate radiological steps. HealthBench \cite{HealthBench} evaluates multi-step clinical tasks spanning triage and critical decision support. AgentClinic \cite{schmidgall2024agentclinic} and SD-Bench \cite{nori2025sequential} simulate dynamic, multi-turn clinical encounters requiring models to actively gather patient information. MedChain \cite{liu2026medchain} provides an interactive environment measuring agent decision-making across sequential patient visits. MedThink-Bench \cite{zhou2025automating} evaluates step-by-step logical traces using expert-annotated rationales. To scale evaluation, automated judging frameworks use clinician-aligned models to score intermediate reasoning paths \cite{zhou2025automating, llmjury2026}. MedRECT \cite{iwase2025medrect} specifically evaluates error correction capabilities within clinical text reasoning.

\paragraph{Visual Interpretability Assessment} Frameworks increasingly evaluate whether vision-language models provide spatial visual evidence supporting their conclusions. PRS-Med \cite{trinh2025prsmed} combines vision-language features with image segmentation to output precise anatomical masks alongside diagnostic text. RJUA-MedDQA \cite{RJUA-MedDQA} tests whether models correctly extract information from medical documents and cite relevant supporting context. GEMeX \cite{GEMeX, GEMeX-ThinkVG} and AOR \cite{AOR} evaluate region-level, multi-step visual reasoning on chest radiographs.

\section{Discussion}
\label{sec:discussion}

Key challenges in reasoning reliability, multimodal fusion, efficiency, and clinical integration must be addressed before deploying medical reasoning models in practice.

\paragraph{The Faithfulness-Plausibility Gap} Plausible hallucinations—generating convincing but medically incorrect rationale paths—present serious safety risks \cite{Reasonmed, HuatuoGPT-o1, soffer2025pitfalls}. Mitigating these errors requires enforcing explicit structural constraints, such as restricting reasoning paths to validated knowledge graphs \cite{MedReason} or decision trees \cite{zang2025medical}. Uncertainty calibration techniques force models to report lower confidence when logic steps diverge \cite{chen2024cod}. Adaptive $\alpha$--$\beta$ frameworks optimize the balance between rationale consistency and novelty under patient distribution shifts to suppress plausible hallucinations \cite{alpha_beta_pncd_2026}. Explicit uncertainty estimation via iterative sampling helps identify factual errors, converting statistical generation into reliable cognitive support where human clinicians maintain analytical oversight \cite{yadkori2024believe, kim2024epistemology}. To bridge this gap, future architectures must combine process-level self-consistency verification with real-time logical validation, allowing models to detect epistemic boundaries and refrain from ungrounded speculation before outputting final clinical recommendations.

Epistemological analyses indicate LLMs lack integrated conceptual knowledge structures, frequently failing to separate primary clinical principles from secondary factors \cite{kim2024epistemology}. In high-stakes settings, models can exhibit overconfidence and fail to abstain when evidence is insufficient \cite{atif2025sacred}. Consequently, LLMs function effectively as cognitive analytical tools rather than autonomous clinical decision-makers \cite{kim2024epistemology}. Trustworthy clinical deployment requires explicit operational boundaries, robust abstention mechanisms, and uncertainty-aware safety guardrails.

\paragraph{Toward Native Multimodal Reasoning} Current multimodal vision-language models often rely on late-fusion architectures that combine visual and textual representations late in processing \cite{pan2025medvlm, Med-R1, Med-RLVR}. Moving toward native multimodal architectures requires interleaving image patches and text tokens within a unified, continuous representation space to avoid the information losses of late-fusion pipelines \cite{wang2024emu3, team2026kimi}. Early token-level integration supports continuous visual token inference, bypassing text-only limitations to enable precise anatomical visual grounding, lesion tracking, and image-based counterfactual analysis \cite{qin2025chain, wu2026lavit}. Systems like Emu3 \cite{wang2024emu3} and Kimi 2.5 \cite{team2026kimi} process visual and textual tokens jointly. Latent visual reasoning strategies \cite{qin2025chain, wu2026lavit} optimize internal steps within continuous visual spaces, advancing reliable multimodal reasoning.

\paragraph{The Efficiency-Performance Frontier} Multi-path reasoning techniques \cite{wei2023chainofthoughtpromptingelicitsreasoning, wang2023self, tang2024medagents, hong2024argmed} require substantial compute resources and increase generation latency. To improve operational efficiency, modern architectures move away from brute-force sampling toward lightweight post-hoc rankers \cite{shi2024medadapter} or meta-cognitive inference managers that dynamically control test-time compute density \cite{medcog2026}. Furthermore, applying targeted reinforcement learning to smaller base models enables concise causal reasoning paths to emerge, reducing latency during time-sensitive clinical operations \cite{zhang2025medtvt, liu2025beyond}. Empirical evaluations \cite{jmir2026deepseek, frontiers2026local} demonstrate that optimizing reasoning trajectories remains essential for acute medical scenarios \cite{frontiers2026local, frontiers2026er} regardless of model parameter scale.

\paragraph{Evaluation Beyond Task Accuracy} Over-reliance on static answer accuracy creates an evaluation gap. Realistic validation requires dynamic, multi-turn benchmarks that simulate longitudinal patient care across multimodal data \cite{AOR, GEMeX}. Addressing benchmark saturation requires adopting interactive simulation environments that mirror real-world patient progression and clinical workflow constraints \cite{HealthBench, schmidgall2024agentclinic, nori2025sequential}. Process-level evaluation must leverage clinician-aligned automated judges to evaluate factual correctness, safety thresholds, and guideline compliance across entire patient care trajectories \cite{zhou2025automating, llmjury2026}. Multimodal benchmarks must also incorporate quantitative visual grounding and interpretability metrics.

\paragraph{Prerequisites for Responsible Clinical Adoption} Safe deployment requires establishing a foundation of sociotechnical reliability (Appendix \ref{sec:appendix_responsible}). Compliance with patient privacy laws (e.g., HIPAA, GDPR) makes privacy-preserving techniques like Federated Learning (FL) essential \cite{jahan2025federated, abbas2024comprehensive, zhao2025privacy}. Deployment requires actively auditing algorithmic bias, mitigating demographic disparities, and addressing cognitive biases in training data \cite{kim2025llm}. Establishing clinical trust involves addressing liability gaps \cite{habli2020artificial} through defined governance frameworks \cite{iti2024accountability}, human-centered explainable AI \cite{thirunavukarasu2023large, hong2024argmed}, and Human-in-the-Loop workflows managed under agile regulatory oversight \cite{USFDA2024PCCPFinal}.

Transparent user interfaces are indispensable in clinical environments to support legal and clinical accountability \cite{10.3389/frai.2022.879603}. In high-acuity settings, clinicians remain legally and ethically responsible for patient care and cannot accept unverified recommendations from black-box systems. Opaque outputs prevent clinicians from evaluating causal logic, creating operational risks. To foster appropriate trust, interfaces must incorporate interactive dialogue and guided deferral protocols by displaying visual attention overlays on medical images or tabular data, generating traceable citations to established clinical guidelines, and highlighting internal model uncertainty during multi-step reasoning. When a model proposes a diagnostic trajectory, the interface must allow clinicians to query individual intermediate logic steps, test counterfactual scenarios, and inspect underlying database queries \cite{electronics13112002, strong2024humanaicollaborationhealthcareguided}. Converting opaque statistical probabilities into auditable cognitive trails enables clinicians to verify clinical reasoning and incorporate AI insights safely into daily care workflows \cite{trust_ai_2026}.

\section{Conclusion}
This review surveyed the development of specialized LLMs for medical reasoning, presenting a technical dual taxonomy spanning training-time and test-time enhancement techniques. We analyzed these methods across data modalities and clinical workflows, tracing the paradigm shift from static answer accuracy to process-level reasoning evaluation. Key remaining challenges include model faithfulness, native multimodal integration, compute efficiency, and safe sociotechnical integration. Addressing these operational barriers will enable medical LLMs to serve as verifiable, trustworthy decision-support tools in modern healthcare.

\section*{Limitations}
Several limitations of this review should be noted. The scope is restricted to publicly available academic literature, intentionally excluding proprietary commercial models whose internal architectural details are undisclosed. Although systematic search strings were used, some relevant studies using non-standard terminology may have been omitted. Furthermore, our proposed taxonomy reflects a specific categorization perspective, and alternative structural frameworks could be formulated. Finally, prioritizing broad panoramic coverage limited deep technical analysis of every individual method. Given the rapid pace of advances in medical AI, this review serves as a snapshot of current progress, and continuous updates will be required.

\bibliography{custom}
\clearpage
\appendix
\section*{Appendix}

This appendix provides supplementary material supporting the main text, including extended background on medical LLMs (Appendix \ref{app:background}), taxonomy visual representations (Appendix \ref{app:taxonomy_visuals}), clinical cognitive stage mappings (Appendix \ref{app:cognitive_stages}), a Methodological Preference Matrix (Appendix \ref{app:methodological_matrix}), detailed technical analysis of knowledge-enhanced reasoning (Appendix \ref{app:knowledge_reasoning}), extended discussions on responsible deployment and model bias (Appendix \ref{sec:appendix_responsible} and \ref{sec:appendix_bias}), an analytical comparison of evaluation benchmarks (Appendix \ref{app:benchmark_comparison}), explanations for benchmark transfer gaps (Appendix \ref{app:transfer_gaps}), and detailed model benchmark performance tables (Appendix \ref{app:performance_tables}).

\section{Background} 
\label{app:background}

Medical reasoning is central to professional clinical practice. While general LLMs excel at processing clinical narrative text, standard probabilistic sequence models are not natively optimized for structured, multi-step inference required in clinical decision-making. This limitation has motivated the development of specialized reasoning LLMs designed to generate verifiable, transparent, and robust diagnostic pathways \cite{savage2024diagnostic, kim2024mdagents, FineMedLM-o1}.

\subsection{From Generalist LLMs to Specialized Reasoners}
Standard probabilistic sequence models excel at pattern retrieval but often fail when executing formal logical operations, such as distinguishing correlation from causation or managing diagnostic uncertainty \cite{wei2023chainofthoughtpromptingelicitsreasoning}. Chain-of-Thought prompting established that requesting step-by-step logic outputs elicits latent reasoning performance \cite{wei2023chainofthoughtpromptingelicitsreasoning, nachane2024few}. Advanced reasoning architectures integrate supervised fine-tuning on structured deductive traces with reinforcement learning from human or AI feedback (RLHF/RLAIF) to systematically reinforce logically sound reasoning steps \cite{openai2024openaio1card, pan2025survey, sonoda2025structured}. Consequently, specialized medical reasoners are defined by their capacity for explicit, auditable logic generation \cite{MedReason}.

\subsection{The Imperative for Robust Reasoning in Medical Practice}
The need for robust medical reasoning is critical because diagnostic errors directly compromise patient safety. Diagnostic inaccuracies account for approximately 31\% of paid medical malpractice claims in the United States \cite{Rates}. Reasoning LLMs offer a structured approach to reduce these errors. In medical education, AI tutoring systems provide interactive feedback to improve student clinical reasoning and history-taking skills \cite{llmimprovedecision}. For practicing physicians, specialized systems analyze clinical documentation in EHRs to identify potential cognitive biases or logical inconsistencies in real time \cite{llmsupport}.

\section{Taxonomy and Framework Visualizations}
\label{app:taxonomy_visuals}

Figure~\ref{fig:taxonomy_appendix} outlines the taxonomy dimensions across data modalities, enhancement methods, clinical applications, and evaluation paradigms. Table~\ref{tab:survey_comparison} compares our scope against existing literature surveys.

\begin{figure*}[htbp]
\centering
\resizebox{\textwidth}{!}{%
\begin{tikzpicture}[
    node distance=0.04cm and 0.18cm,
    font=\sffamily,
    cat1node/.style={rectangle, draw=borderblue, fill=categoryblue!80, font=\bfseries\scriptsize, rounded corners=2pt, inner sep=3.5pt, align=center, text width=1.3cm, minimum width=1.3cm},
    cat15node/.style={rectangle, draw=gray!60, fill=categorygreen!50, font=\bfseries\tiny, rounded corners=2pt, inner sep=2.5pt, align=center, text width=1.1cm, minimum width=1.1cm},
    cat2node/.style={rectangle, draw=gray!50, fill=gray!12, font=\bfseries\tiny, rounded corners=2pt, inner sep=2.5pt, align=center, text width=1.4cm, minimum width=1.4cm},
    itemnode/.style={rectangle, draw=gray!30, fill=white, font=\fontsize{6.5pt}{7.8pt}\selectfont, rounded corners=2pt, inner sep=3.5pt, align=left, text width=7.8cm}
]

\node[itemnode] (r1) {MedReason \cite{MedReason}, EMULATION \cite{sft/EMULATION}, BP4ER \cite{he2024bp4er}, CLINICR \cite{nachane2024few}, MedEx \cite{zafar2025medex}, Med-RLVR \cite{Med-RLVR}, AlphaMed \cite{liu2025beyond}, m1 \cite{huang2025m1}, FineMedLM-o1 \cite{FineMedLM-o1}, MedAdapter \cite{shi2024medadapter}};

\node[itemnode, below=of r1.south west, anchor=north west] (r2) {MC-CoT \cite{wei2024mc}, MedE2 \cite{mu2025elicit}, PRS-Med \cite{trinh2025prsmed}, AOR \cite{AOR}, GEMeX \cite{GEMeX}, MedGround-R1 \cite{MedGroundR1}, MedVLM-R1 \cite{pan2025medvlm}, MedTVT-R1 \cite{zhang2025medtvt}};

\node[itemnode, below=of r2.south west, anchor=north west] (r3) {MedAgentGym \cite{xu2025medagentgym}, BioCoder \cite{tang2023biocoder}, BioDSBench \cite{wang2024can}, EHRAgent \cite{shi2024ehragentcodeempowerslarge}};

\node[itemnode, below=0.10cm of r3.south west, anchor=north west] (r4) {\textbf{Multi-stage}: FineMedLM-o1 \cite{FineMedLM-o1}, Clin-RaGen \cite{ClinRaGen}, AOR \cite{AOR}; \textbf{Chain-aware}: Med-PaLM 2 \cite{Medpalm2}, HuatuoGPT-o1 \cite{HuatuoGPT-o1}, EMULATION \cite{sft/EMULATION}, BioMed-R1 \cite{sft/BioMed-R1}, MedReason \cite{MedReason}; \textbf{Position-aware}: PRS-Med \cite{trinh2025prsmed}, MedGround-R1 \cite{MedGroundR1}, AOR \cite{AOR}};

\node[itemnode, below=of r4.south west, anchor=north west] (r5) {\textbf{RLHF}: Med-PaLM 2 \cite{Medpalm2}; \textbf{RLAIF}: HuatuoGPT-o1 \cite{HuatuoGPT-o1}; \textbf{Structured/GRPO}: MedTVT-R1 \cite{zhang2025medtvt}, Clin-RaGen \cite{ClinRaGen}, BioMed-R1 \cite{sft/BioMed-R1}, MedGround-R1 \cite{MedGroundR1}, AlphaMed \cite{liu2025beyond}, Med-R1 \cite{Med-R1}, GMAI-VL-R1 \cite{GMAI-VL-R1}};

\node[itemnode, below=of r5.south west, anchor=north west] (r6) {Clinical CoT \cite{kwon2023large}, DR-CoT \cite{wu2024large}, CoD \cite{chen2024cod}, BP4ER \cite{he2024bp4er}, MC-CoT \cite{wei2024mc}, PathCoT \cite{zhou2025pathcot}, Structured Clinical \cite{structured_clinical_2026}, Evidence-Driven CoT \cite{evidence_driven_cot_2026}};

\node[itemnode, below=of r6.south west, anchor=north west] (r7) {Self-Consistency \cite{wang2023self}, MedAdapter \cite{shi2024medadapter}, Ensemble Reasoning \cite{lucas2024reasoning}, o1 Replication \cite{huang2025o1replicationjourney}, MedCog \cite{medcog2026}, MedPRM \cite{medprm2025}};

\node[itemnode, below=of r7.south west, anchor=north west] (r8) {\textbf{Unstructured}: SelfRewardRAG \cite{selfrewardrag2024}, Self-BioRAG \cite{jeong2024improving}, RARoK \cite{zhan2024rarok}; \textbf{Structured}: ICP \cite{wu2024guiding}, MedReason \cite{MedReason}, MedEx \cite{zafar2025medex}, MedAI+ \cite{medai_plus_2026}};

\node[itemnode, below=of r8.south west, anchor=north west] (r9) {\textbf{Collaborative Deliberation}: ArgMed-Agents \cite{hong2024argmed}, AMANDA \cite{amanda2025}, MultiMedRes \cite{inquire_interact_integrate_2024}; \textbf{Workflow Execution}: MedRAX \cite{fallahpour2025medrax}, MedChain \cite{liu2026medchain}, ClinicalAgent \cite{yue2024clinicalagentclinicaltrialmultiagent}, MedAgentGym \cite{xu2025medagentgym}; \textbf{Quality Evaluation}: Medical AI Consensus \cite{radiologyMA2025}};

\node[itemnode, below=0.10cm of r9.south west, anchor=north west] (r10) {EHRAgent \cite{shi2024ehragentcodeempowerslarge}, Med-Copilot \cite{xu2025medagentgym}, AIPatient \cite{yu2024aipatientsimulatingpatientsehrs}, ClinicalLab \cite{yan2024clinicallabaligningagentsmultidepartmental}};

\node[itemnode, below=of r10.south west, anchor=north west] (r11) {HuatuoGPT-o1 \cite{HuatuoGPT-o1}, MedReason \cite{MedReason}, ReasonMed \cite{Reasonmed}, Dual-Inf \cite{zhou2025dualinf}, FineMedLM-o1 \cite{FineMedLM-o1}, PRS-Med \cite{trinh2025prsmed}};

\node[itemnode, below=of r11.south west, anchor=north west] (r12) {MedWorldModel \cite{MedWorldModel}, GPT-RadPlan \cite{GPT-RadPlan}, MedRBench \cite{MedRBench}};

\node[itemnode, below=of r12.south west, anchor=north west] (r13) {CardioCoT \cite{CardioCoT}, RiskPredict-LLM \cite{riskpredict_llm}};

\node[itemnode, below=of r13.south west, anchor=north west] (r14) {DrugAgent \cite{liu2024drugagent}, ChemCrow \cite{ChemCrow}, AgentMol \cite{karabowicz2025agentmol}};

\node[itemnode, below=0.10cm of r14.south west, anchor=north west] (r15) {MedQA \cite{jin2020diseasedoespatienthave}, PubMedQA \cite{jin2019pubmedqa}, MedXpertQA \cite{medxpertqa}, MedAgentsBench \cite{medagentsbench}, MR-Bench \cite{ren2026mrbench}, ER-Reason \cite{mehandru2026erreason}, mARC \cite{shidara2026marc}};

\node[itemnode, below=of r15.south west, anchor=north west] (r16) {MedR-Bench \cite{MedRBench}, RadRScore \cite{ChestX-Reasoner}, HealthBench \cite{HealthBench}, AgentClinic \cite{schmidgall2024agentclinic}, SD-Bench \cite{nori2025sequential}, MedChain \cite{liu2026medchain}, MedThink-Bench \cite{zhou2025automating}};

\node[itemnode, below=of r16.south west, anchor=north west] (r17) {PRS-Med \cite{trinh2025prsmed}, RJUA-MedDQA \cite{RJUA-MedDQA}, GEMeX \cite{GEMeX}, AOR \cite{AOR}, GEMeX-ThinkVG \cite{GEMeX-ThinkVG}};

\node[cat2node, left=0.18cm of r1.west, anchor=east] (c2_1) {Text};
\node[cat2node, left=0.18cm of r2.west, anchor=east] (c2_2) {Vision};
\node[cat2node, left=0.18cm of r3.west, anchor=east] (c2_3) {Code};

\node[cat2node, left=0.18cm of r4.west, anchor=east] (c2_4) {Supervised Fine-Tuning};
\node[cat2node, left=0.18cm of r5.west, anchor=east] (c2_5) {Reinforcement Learning};
\node[cat2node, left=0.18cm of r6.west, anchor=east] (c2_6) {Prompting};
\node[cat2node, left=0.18cm of r7.west, anchor=east] (c2_7) {Selection \& Aggregation};
\node[cat2node, left=0.18cm of r8.west, anchor=east] (c2_8) {Knowledge-Enhanced};
\node[cat2node, left=0.18cm of r9.west, anchor=east] (c2_9) {Multi-Agent Systems};

\node[cat2node, left=0.18cm of r10.west, anchor=east] (c2_10) {EHR \& History};
\node[cat2node, left=0.18cm of r11.west, anchor=east] (c2_11) {Diagnosis \& DDx};
\node[cat2node, left=0.18cm of r12.west, anchor=east] (c2_12) {Treatment Planning};
\node[cat2node, left=0.18cm of r13.west, anchor=east] (c2_13) {Prognosis \& Risk};
\node[cat2node, left=0.18cm of r14.west, anchor=east] (c2_14) {Drug Discovery};

\node[cat2node, left=0.18cm of r15.west, anchor=east] (c2_15) {Answer Accuracy};
\node[cat2node, left=0.18cm of r16.west, anchor=east] (c2_16) {Reasoning Quality};
\node[cat2node, left=0.18cm of r17.west, anchor=east] (c2_17) {Visual Interpretability};

\node[cat15node, left=0.18cm of $(c2_4.west)!0.5!(c2_5.west)$, anchor=east] (trn_node) {Training-time};
\node[cat15node, left=0.18cm of $(c2_6.west)!0.5!(c2_9.west)$, anchor=east] (tst_node) {Test-time};

\coordinate (c1_x) at ($(trn_node.west)+(-0.30cm, 0)$);

\node[cat1node, anchor=east] (c1_mod) at (c1_x |- {$(c2_1.north)!0.5!(c2_3.south)$}) {\textbf{Modality}};

\node[cat1node, anchor=east] (c1_met) at (c1_x |- {$(trn_node.north)!0.5!(tst_node.south)$}) {\textbf{Methods}};

\node[cat1node, anchor=east] (c1_app) at (c1_x |- {$(c2_10.north)!0.5!(c2_14.south)$}) {\textbf{Applications}};

\node[cat1node, anchor=east] (c1_eval) at (c1_x |- {$(c2_15.north)!0.5!(c2_17.south)$}) {\textbf{Evaluation}};

\foreach \i in {1,2,3} {
    \draw[->, gray!60] (c1_mod.east) -- ++(0.08,0) |- (c2_\i.west);
    \draw[->, gray!60] (c2_\i.east) -- (r\i.west);
}

\draw[->, gray!60] (c1_met.east) -- ++(0.08,0) |- (trn_node.west);
\draw[->, gray!60] (c1_met.east) -- ++(0.08,0) |- (tst_node.west);

\draw[->, gray!60] (trn_node.east) -- ++(0.08,0) |- (c2_4.west);
\draw[->, gray!60] (trn_node.east) -- ++(0.08,0) |- (c2_5.west);

\foreach \i in {6,7,8,9} {
    \draw[->, gray!60] (tst_node.east) -- ++(0.08,0) |- (c2_\i.west);
}
\foreach \i in {4,5,6,7,8,9} {
    \draw[->, gray!60] (c2_\i.east) -- (r\i.west);
}

\foreach \i in {10,11,12,13,14} {
    \draw[->, gray!60] (c1_app.east) -- ++(0.08,0) |- (c2_\i.west);
    \draw[->, gray!60] (c2_\i.east) -- (r\i.west);
}

\foreach \i in {15,16,17} {
    \draw[->, gray!60] (c1_eval.east) -- ++(0.08,0) |- (c2_\i.west);
    \draw[->, gray!60] (c2_\i.east) -- (r\i.west);
}

\end{tikzpicture}%
}
\caption{A taxonomy of medical reasoning with LLMs. This figure provides a visual summary of the topics discussed in this review, outlining the primary data modalities (Section \ref{sec:modalities}), the core reasoning enhancement techniques (Section \ref{sec:architectures}), key medical applications (Section \ref{sec:applications}), and the evolving paradigms for evaluation (Section \ref{sec:evaluation}).}
\label{fig:taxonomy_appendix}
\end{figure*}

\begin{table*}[htbp]
\centering
\caption{Taxonomy Dimension Matrix (Survey Gap Analysis against Existing Literature Reviews).}
\label{tab:survey_comparison}
\small
\setlength{\tabcolsep}{5pt}
\renewcommand{\arraystretch}{1.15}

\begin{tabular}{
    >{\raggedright\arraybackslash}p{0.18\textwidth}
    >{\raggedright\arraybackslash}p{0.36\textwidth}
    >{\raggedright\arraybackslash}p{0.38\textwidth}}
\toprule
\textbf{Survey} &
\textbf{Main Focus} &
\textbf{Difference / Gap Addressed by Ours} \\
\midrule

Zhou et al. \cite{zhou2023survey} &
Reviews the development and deployment of medical LLMs, including model architectures, training data, task performance, applications, and deployment challenges. &
Focuses broadly on medical LLM development and application rather than constructing a taxonomy specifically for medical reasoning enhancement. \\

Shool et al. \cite{shool2025systematic} &
Systematically analyzes evaluation metrics, study designs, application domains, and model distributions in clinical LLM studies. &
Focuses on how medical LLMs are evaluated, whereas our survey examines how medical reasoning capabilities are enhanced and assessed. \\

Berger et al. \cite{berger2025reasoning} &
Reviews CoT prompting, reinforcement learning, specialized reasoning models, prompt optimization, multi-agent systems, and evaluation. &
Mainly organizes the literature around individual techniques and models, while our survey provides a unified training-time versus test-time taxonomy across text, image, and code. \\

Mansoor et al. \cite{mansoor2026reasoning} &
Surveys medical reasoning techniques, model adaptation, multimodal processing, clinical applications, ethical challenges, and clinical integration. &
Places greater emphasis on clinical integration, while our survey focuses on the relationships among reasoning enhancement, selection, verification, and process-oriented evaluation. \\

Peng et al. \cite{peng2025aligning} &
Connects clinical requirements with computational reasoning through Miller's Pyramid and deductive, inductive, and abductive reasoning. &
Focuses on the competencies and logical forms required for medical reasoning, whereas our survey examines how these capabilities are enhanced during training and inference. \\

Lin et al. \cite{lin2026survey} &
Reviews reasoning datasets and training-time and test-time methods, emphasizing the transition from individual models to collaborative agents. &
Uses the evolution toward multi-agent systems as its main perspective, while our survey covers a broader range of mechanisms, modalities, applications, and evaluation paradigms. \\

Ren et al. \cite{ren2026mrbench} &
Organizes medical reasoning into seven technical routes and introduces a unified evaluation and MR-Bench. &
Emphasizes cognitive theories and empirical benchmarking, while our survey more broadly covers text, image, and code reasoning, knowledge conflicts, and deployment challenges. \\

\textbf{Ours} &
\textbf{Establishes a dual taxonomy (training-time vs. test-time) across text, vision, and code modalities; aligns techniques with clinical cognitive stages; analyzes dynamic knowledge conflict resolution and process-level evaluation.} &
- \\

\bottomrule
\end{tabular}
\end{table*}

\section{Cognitive Stages of Clinical Reasoning}
\label{app:cognitive_stages}

Table~\ref{tab:cognitive_stages} maps specific LLM mechanisms onto the cognitive stages of clinical practice.

\begin{table*}[htbp]
\centering
\caption{Taxonomy of Medical Reasoning Techniques based on Clinical Cognitive Stages.}
\label{tab:cognitive_stages}
\small
\setlength{\tabcolsep}{5pt}
\renewcommand{\arraystretch}{1.2}
\begin{tabularx}{\textwidth}{@{}p{3.0cm} X p{4.2cm}@{}}
\toprule
\textbf{Division} & \textbf{Definition} & \textbf{Representative Frameworks \& Models} \\
\midrule
Hypothesis Generation \& Differential Diagnosis \cite{kwon2023large, tang2024medagents} & Triggered by initial problem representation, this phase involves formulating a prioritized list of potential diseases. Corresponding LLM techniques activate relevant illness scripts and generate a wide range of potential answers. & Clinical CoT \cite{kwon2023large}, PathCoT \cite{zhou2025pathcot}, Dr. Knows \cite{gao2025leveraging}, FineMedLM-o1 \cite{FineMedLM-o1}, Disc-MedLLM \cite{bao2023discmedllm}, MedAgents \cite{tang2024medagents} \\
\midrule
Evidence Gathering \& Hypothesis Refinement \cite{he2024bp4er, zafar2025medex} & An iterative, cyclical process where new information is sought to discriminate between competing hypotheses. Models in this category focus on targeted information retrieval (RAG) and Chain-of-Thought (CoT) methods to guide sequential inquiry. & BP4ER \cite{he2024bp4er}, MedEx \cite{zafar2025medex}, ChestX-Reasoner \cite{ChestX-Reasoner}, Med-RLVR \cite{Med-RLVR}, EHRAgent \cite{shi2024ehragentcodeempowerslarge}, MedAgentGym \cite{xu2025medagentgym} \\
\midrule
Synthesis \& Verification \cite{AOR, Reasonmed} & All accumulated evidence is integrated to form a coherent, verified narrative explaining the patient's condition. This involves higher-order cognitive skills and integrating multi-source evidence to confirm the final working diagnosis. & AOR \cite{AOR}, MedGround-R1 \cite{MedGroundR1}, ReasonMed \cite{Reasonmed}, GEMeX \cite{GEMeX}, GEMeX-ThinkVG \cite{GEMeX-ThinkVG} \\
\bottomrule
\end{tabularx}
\end{table*}

\paragraph{Hypothesis Generation \& Differential Diagnosis} Clinical CoT \cite{kwon2023large} and PathCoT \cite{zhou2025pathcot} use prompting to externalize hypothesis generation. MedAgents \cite{tang2024medagents} deploys multi-agent debate to propose differential diagnoses, while Dr. Knows \cite{gao2025leveraging} grounds initial diagnostic hypotheses in knowledge graphs. FineMedLM-o1 \cite{FineMedLM-o1} optimizes hypothesis ranking via SFT, and Disc-MedLLM \cite{bao2023discmedllm} tests differential lists against clinical confounders.

\paragraph{Evidence Gathering \& Hypothesis Refinement} BP4ER \cite{he2024bp4er} structures evidence collection using least-to-most prompting. MedEx \cite{zafar2025medex} integrates iterative inquiry with First-Order Logic to evaluate diagnostic assertions. ChestX-Reasoner \cite{ChestX-Reasoner} performs step-by-step verification of radiological findings. Med-RLVR \cite{Med-RLVR} applies reinforcement learning to reward correct evidence interpretation, while BioCoder \cite{tang2023biocoder}, MedAgentGym \cite{xu2025medagentgym}, and BioDSBench \cite{wang2024can} execute code-based queries to collect structured clinical evidence.

\paragraph{Synthesis \& Verification} AOR \cite{AOR} achieves semantic evidence synthesis through anatomical ontology alignment. MedGround-R1 \cite{MedGroundR1} integrates spatial bounding boxes to anchor text assertions in visual findings. ReasonMed \cite{Reasonmed} generates auditable reasoning chains for diagnostic verification, while GEMeX \cite{GEMeX} and GEMeX-ThinkVG \cite{GEMeX-ThinkVG} perform region-level visual rationale evaluation.

\section{Methodological Preference Matrix}
\label{app:methodological_matrix}

Table~\ref{tab:methodological_matrix} presents a comparative preference matrix cross-referencing technical subcategories against their core assumptions, failure modes, deployment constraints, and preferred clinical applications.

\begin{table*}[htbp]
\centering
\caption{Methodological Preference Matrix of Medical Reasoning Enhancement Paradigms.}
\label{tab:methodological_matrix}
\fontsize{7.5pt}{8.8pt}\selectfont
\setlength{\tabcolsep}{4pt}
\renewcommand{\arraystretch}{1.15}
\begin{tabularx}{\textwidth}{@{}p{2.2cm} p{3.0cm} p{3.2cm} p{2.8cm} X@{}}
\toprule
\textbf{Method Class} & \textbf{Core Technical Assumptions} & \textbf{Clinical / Logical Failure Modes} & \textbf{Clinical Deployment Constraints} & \textbf{Preferred Clinical Scenario} \\
\midrule
\textbf{Multi-Stage SFT} & Task decomposition into structured fact-to-abstraction layers builds parameter stability. & Risk of upstream alignment errors propagating into secondary downstream inference levels. & Requires highly structured data hierarchies; compute scales linearly with training data. & Multimodal imaging scenarios where basic structural identification precedes diagnostic extraction. \\
\midrule
\textbf{Chain-Aware SFT} & Tuning internal parameters directly on explicit reasoning traces transfers procedural logic. & Stylistic emulation where models replicate linguistic templates without mastering causality. & High data engineering overhead; heavy reliance on expert tokens or teacher outputs. & Automated compilation of trace-auditable differential diagnoses from patient profiles. \\
\midrule
\textbf{Position-Aware SFT} & Mandating strict token-to-region or token-to-mask localization enforces spatial interpretability. & Attention drift where language generation segments diverge from targeted anatomical bounding boxes. & Dependent on pixel-level clinical masks; restricted architecture scaling. & High-precision lesion mapping and automated radiography report compilation. \\
\midrule
\textbf{RLHF / RLAIF} & Adjusting baseline parameter spaces via curated preference vectors ensures clinical safety. & Risk of over-conservatism, leading to high token refusal rates on legitimate edge cases. & High compute overhead for reward network tracking; expert evaluation bottlenecks. & Conversational patient triage systems and patient-facing interface interactions. \\
\midrule
\textbf{Structured Reward RL (GRPO)} & Mathematical optimization over composite reward criteria forces implicit reasoning paths to form. & Reward-hacking tendencies where paths comply with strict format structures over absolute truth. & High parallel generation and sampling compute constraints during policy updates. & Verifiable ECG classification, multi-step math logic, and tightly rule-gated clinical QA. \\
\midrule
\textbf{Prompt-Based Elicitation} & Domain-specific input framing and clinical workflows activate internal paths without weight changes. & Vulnerable to extreme variance across sampling runs; high logical hallucination rates. & Inbound context window bottlenecks; high test-time prompt optimization requirements. & Low-compute clinical endpoints requiring instantaneous cold-start symptom parsing. \\
\midrule
\textbf{Reasoning Selection \& Aggregation} & Sampling decoding paths yields a consensus distribution filtering out random generation errors. & Diminishing returns if base networks suffer from embedded structural biases or shared hallucinations. & Severe computational latency footprints due to massive token generation requirements. & Critical diagnostic verification checks where processing latency is subordinate to logic accuracy. \\
\midrule
\textbf{Unstructured Knowledge RAG} & Injecting real-time contexts from vast medical databases compensates for parametric knowledge drift. & Generation distraction and loss of focus caused by noisy or irrelevant data fragments. & Requires real-time vector database query scaling; pipeline query latency overhead. & Longitudinal profile aggregation and integration of recently updated healthcare guidelines. \\
\midrule
\textbf{Structured Knowledge (KGs / FOL)} & Constraining logical transitions along semantic triples or symbolic rules eliminates physiological hallucinations. & Complete failure when encountering out-of-vocabulary terms; fragile under imperfect graph data. & Severe database lookup bottlenecks; graph scaling and ontology tracking maintenance costs. & Multi-step medical calculator execution and strict regulatory compliance verification. \\
\midrule
\textbf{Collaborative Deliberation Agents} & Separating logic tracking into specialized debate, critique, and proposer nodes drives path optimization. & Risk of unconstructive debate loops and deadlocks when managing highly ambiguous signals. & High inter-agent message orchestration costs; exponential API or hardware compute tracking costs. & Complex zero-shot cross-specialty consultations and open-ended clinical deliberations. \\
\midrule
\textbf{Workflow Execution Agents} & Delegating execution branches to tool-using, code-generating modules protects against task collapse. & Cascading modular failure loops where early tool exceptions destabilize subsequent pipeline stages. & Requires rigid sandbox execution environments; high systems orchestration complexity. & Automated clinical trial configuration design and multi-department medical record reasoning. \\
\midrule
\textbf{Quality Evaluation Agents} & Isolating critic parameters from generation components yields stable, human-aligned sequence scoring. & Criterion drift behaviors where judging matrices favor stylistic fluency over clinical truth. & High continuous inference resource footprints during runtime quality checking routines. & Automated sequence validation checking, sandbox output scoring, and logic path termination. \\
\bottomrule
\end{tabularx}
\end{table*}

\section{A Deeper Dive into Knowledge-Enhanced Reasoning}
\label{app:knowledge_reasoning}

\subsection{Methodological Categorization}
\paragraph{Unstructured Knowledge Contextualization} Standard Retrieval-Augmented Generation (RAG) fetches relevant literature or EHR text at query time to ground model answers \cite{jeong2024improving, zhan2024rarok}. Advanced implementations incorporate multi-step iterative retrieval \cite{yu2023chain} or self-evaluation loops to prune non-informative context \cite{selfrewardrag2024}.

\paragraph{Structured Knowledge Guidance} Uses KGs and ontologies to constrain generation: (1) \textit{Constrained Generation} (inserting structured UMLS triplets into contexts \cite{wu2024guiding}); (2) \textit{Path Validation} (checking intermediate reasoning steps against KG paths); (3) \textit{Hybrid Integration} (combining text retrieval with KG graph traversals \cite{cabello2024meg}); and (4) \textit{Formal Symbolic Rules} (applying First-Order Logic constraints \cite{zafar2025medex}).

\subsection{Conflict Resolution in Knowledge Integration}
\label{sec:conflict_resolution_appendix}
Knowledge integration must handle contradictions between static medical ontologies and dynamic retrieved literature. In our asymmetric evidence framework, established medical ontologies act as non-negotiable safety guardrails. Dynamic text overrides static KGs only when supported by verified, high-impact literature updates. NLI classifiers compute contradiction scores $P(\text{contradiction} \mid \text{KG}, \text{RAG})$ \cite{xu2025mega, zheng2025kcr}. If conflicts cannot be resolved automatically, the system outputs uncertainty flags and triggers human deferral protocols \cite{jullien2024semeval}.

\section{Extended Discussion on Responsible Clinical Adoption}
\label{sec:appendix_responsible}

\paragraph{Addressing Algorithmic and Cognitive Bias Risks} Benchmark evaluations on BiasMedQA \cite{kim2025llm} show that LLMs can replicate human cognitive errors, such as anchoring bias. Mitigation strategies must target underlying reasoning processes rather than relying solely on post-hoc text filters.

\paragraph{Navigating Regulatory Frameworks and Accountability} Regulatory agencies are developing dynamic oversight frameworks, such as the FDA guidance on Predetermined Change Control Plans (PCCP) \cite{USFDA2024PCCPFinal}, which defines protocols for iterative algorithm updates under pre-specified safety limits.

\paragraph{Technical Approaches to Privacy and Data Security} Federated Learning (FL) enables collaborative model training across clinical institutions without centralizing sensitive patient EHR data \cite{zhao2025privacy, jahan2025federated}.

\paragraph{Responsible Deployment Mitigation Checklist}
(1) \textbf{Performance Monitoring}: Continually audit clinical accuracy and bias drift across active deployments.
(2) \textbf{Clinical Oversight}: Enforce Human-in-the-Loop validation for high-stakes clinical recommendations.
(3) \textbf{Escalation Protocols}: Define standardized deferral paths when model uncertainty is elevated.
(4) \textbf{Privacy Compliance}: Implement FL and differential privacy matching HIPAA and GDPR regulations.
(5) \textbf{Bias Auditing}: Conduct systematic evaluations for demographic disparities and heuristic cognitive errors.
(6) \textbf{Regulatory Alignment}: Align continuous learning pipelines with FDA PCCP requirements.
(7) \textbf{Interface Transparency}: Provide user interfaces displaying visual grounding, guideline citations, and uncertainty indicators.

\section{Model Bias and Equity}
\label{sec:appendix_bias}

Medical reasoning LLMs risk amplifying clinical disparities present in training data \cite{mittermaier2023bias, pfohl2024toolbox, omar2025sociodemographic}. These vulnerabilities manifest across three key areas:

\paragraph{Sociodemographic and Vignette Stereotyping}
When processing identical clinical presentations, LLMs sometimes generate disparate diagnostic recommendations based on explicit or implicit demographic markers (e.g., race, gender, age) \cite{omar2025sociodemographic, pfohl2024toolbox}. This occurs when probabilistic models over-index on historical demographic correlations in training text.

\paragraph{Clinical Cognitive and Heuristic Biases}
Reasoning models can inherit human cognitive heuristics from clinical training text, such as anchoring bias (over-relying on initial symptoms), availability bias (over-diagnosing conditions recently highlighted in training data), and confirmation bias (selecting evidence that fits an initial hypothesis while ignoring contradictory test results) \cite{kim2025llm}.

\paragraph{Equity-Aware Mitigation Strategies}
To promote equitable decision support, deployment frameworks should integrate:
\begin{enumerate}
    \item \textbf{Counterfactual Perturbation Testing}: Systematically evaluating model output stability by swapping demographic attributes in prompts to detect differential reasoning paths.
    \item \textbf{Subgroup-Calibrated Preference Alignment}: Applying subgroup-stratified reward functions during RL optimization to penalize demographic disparities in diagnostic accuracy.
    \item \textbf{Multi-Center Cohort Validation}: Auditing reasoning models across multi-center patient populations representing diverse demographic backgrounds prior to clinical deployment.
\end{enumerate}

\section{Comparative Analysis of Evaluation Benchmarks}
\label{app:benchmark_comparison}

Table~\ref{tab:appendix-benchmark-comparison} compares representative benchmarks across Answer Accuracy, Reasoning Quality, and Visual Interpretability. While static multiple-choice benchmarks (MedQA \cite{jin2020diseasedoespatienthave}, PubMedQA \cite{jin2019pubmedqa}) evaluate factual recall, dynamic benchmarks (HealthBench \cite{HealthBench}, AgentClinic \cite{schmidgall2024agentclinic}, SD-Bench \cite{nori2025sequential}) simulate longitudinal clinical interactions and diagnostic triage.

\begin{table*}[htbp]
\centering
\caption{A comparative analysis of key benchmarks for medical reasoning evaluation.}
\label{tab:appendix-benchmark-comparison}
\small
{
\setlength{\tabcolsep}{3pt}
\newcolumntype{Y}{>{\centering\arraybackslash}X}
\begin{tabularx}{\textwidth}{@{} >{\raggedright}p{1.5cm} >{\raggedright}p{2.2cm} c *{5}{Y} @{}}
\toprule
\textbf{Type} & \textbf{Benchmark} & \textbf{\#Size} & 
{\footnotesize\bfseries\shortstack{Multi-turn \\ Conversation}} & 
{\footnotesize\bfseries\shortstack{Clinical \\ Scenarios}} & 
{\footnotesize\bfseries\shortstack{Diverse \\ Specialties}} & 
{\footnotesize\bfseries\shortstack{Reasoning \\ Step Eval.}} & 
{\footnotesize\bfseries\shortstack{Multi-dim. \\ Eval.}} \\
\midrule
\multirow{4}{=}{\raggedright \textbf{Answer Accuracy}} 
 & MedQA \cite{jin2020diseasedoespatienthave} & 1,273 & $\times$ & \checkmark & $\times$ & $\times$ & $\times$ \\
 & PubMedQA \cite{jin2019pubmedqa} & 1,000 & $\times$ & $\times$ & $\times$ & $\times$ & $\times$ \\
 & MedXpertQA \cite{medxpertqa} & 4,450 & $\times$ & \checkmark & \checkmark & $\times$ & $\times$ \\
 & MedAgentsBench \cite{medagentsbench} & 862 & $\times$ & \checkmark & $\times$ & $\times$ & $\times$ \\ 
 \midrule
\multirow{4}{=}{\raggedright \textbf{Reasoning Quality}} 
 & MedBench \cite{cai2023medbenchlargescalechinesebenchmark} & 1,453 & $\times$ & \checkmark & \checkmark & \checkmark & \checkmark \\
 & HealthBench \cite{HealthBench} & 5,000 & \checkmark & \checkmark & \checkmark & \checkmark & \checkmark \\
 & AgentClinic \cite{schmidgall2024agentclinic} & N/A (Interactive Env) & \checkmark & \checkmark & \checkmark & $\times$ & $\times$ \\
 & SD-Bench \cite{nori2025sequential} & 304 & \checkmark & \checkmark & $\times$ & $\times$ & \checkmark \\ 
 \midrule
\multirow{2}{=}{\raggedright \textbf{Visual Interpretability}} 
 & PRS-Med \cite{trinh2025prsmed} & N/A (Segmentation) & $\times$ & $\times$ & \checkmark & $\times$ & \checkmark \\
 & RJUA-MedDQA \cite{RJUA-MedDQA} & 2,000 & $\times$ & \checkmark & $\times$ & \checkmark & $\times$ \\ 
\bottomrule
\end{tabularx}
}
\end{table*}

\section{Mechanistic Explanations for Transfer Gaps}
\label{app:transfer_gaps}
Performance gains on multiple-choice QA datasets often fail to transfer to practical clinical reasoning due to: (1) \textit{Task Granularity Misalignment} (factual recall vs. multi-step causal deduction \cite{njpdm_eval, chen2026beyond}); (2) \textit{Modal Specificity Gaps} (text-only performance failing in late-fusion vision-language architectures \cite{pan2025medvlm, mu2025elicit}); and (3) \textit{Clinical Pragmatics Deficits} (ignoring incomplete or noisy clinical real-world data \cite{sim2025critique, kim2025limitations}). Proposed mitigations include clinical reasoning curriculum fine-tuning \cite{FineMedLM-o1, AOR}, clinical reward reinforcement learning \cite{Med-RLVR, ClinRaGen}, and knowledge-guided clinical scaffolding \cite{MedReason, zafar2025medex}.

\section{Detailed Performance on Benchmarks}
\label{app:performance_tables}

Tables~\ref{tab:appendix-train-results}, \ref{tab:appendix-test-results}, \ref{tab:mmlu_pro_comparison}, and \ref{tab:mmlu_training_progression} report model benchmark performance from literature. Note: Metrics across rows reflect reported literature values across varying task settings rather than direct head-to-head experimental comparisons.

\begin{table*}[htbp]
\centering
\caption{Evaluation Results of Training-time Enhanced Models (SFT \& RL). Note: Values represent task-isolated literature reporting across varying settings rather than direct horizontal head-to-head comparisons.}
\label{tab:appendix-train-results}
\setlength{\tabcolsep}{6pt}
\renewcommand{\arraystretch}{1.15}
\begin{tabular}{l c c c}
\toprule
\textbf{Model} & \textbf{Size} & \textbf{Dataset} & \textbf{Metric / Acc.} \\
\midrule
FineMedLM \cite{FineMedLM-o1}               & 8B   & MMLU                & 71.4 \\
FineMedLM-o1 \cite{FineMedLM-o1}             & 8B   & MMLU                & 78.2 \\
FineMedLM-o1 (TTT) \cite{FineMedLM-o1}      & 8B   & MMLU                & 81.36 \\
HuatuoGPT-o1 \cite{HuatuoGPT-o1}              & 8B   & MMLU-Pro            & 58.7 \\
HuatuoGPT-o1 \cite{HuatuoGPT-o1}              & 70B  & MMLU-Pro            & 71.0 \\
HuatuoGPT-II \cite{HuatuoGPT-o1}    & 7B  & CMMLU                & 59.1 \\
HuatuoGPT-II \cite{HuatuoGPT-o1}   & 13B   & CMMLU                & 61.5 \\
BioMed-R1 \cite{sft/BioMed-R1}                 & 8B   & MMLU                & 71.1 \\
BioMed-R1 \cite{sft/BioMed-R1}               & 32B  & MMLU                & 80.7 \\
Med-PaLM 2 (ER) \cite{Medpalm2}           & -    & MMLU                & 92.0 \\
Med-PaLM 2 (best) \cite{Medpalm2}         & -    & MMLU                & 92.0 \\
MedReason (LLaMA) \cite{MedReason}          & 8B   & MMLU-Pro            & 63.1 \\
MedReason (Mistral) \cite{MedReason}       & 7B   & MMLU-Pro            & 50.8 \\
AlphaMed \cite{liu2025beyond}                  & 7B   & MMLU                 & 66.8 \\
Clin-RaGen \cite{ClinRaGen}                & 793M & MIMIC-III (Micro)   & 0.6501 (F1) \\
Clin-RaGen \cite{ClinRaGen}                & 793M & MIMIC-IV (Micro)    & 0.7127 (F1) \\
AOR \cite{AOR}                       & -    & CheXpert (closed)   & 71.58 \\
AOR \cite{AOR}                  & -    & CheXpert (open)     & 53.85 \\
MedTVT-R1 \cite{zhang2025medtvt}                 & -    & ECG-QA              & 0.0831 (BLEU) \\
PRS-Med \cite{trinh2025prsmed}                   & -    & Lung X-ray    & 0.942 (mIoU) \\
PRS-Med \cite{trinh2025prsmed}                   & -    & Lung X-ray          & 0.969 (mDice) \\
MedGround-R1 \cite{MedGroundR1}              & -    & MS-CXR              & 83.12 \\
AlphaMed (LLaMA3.1) \cite{liu2025beyond}       & 8B   & MMLU-Pro            & 59.3 \\
Med-R1 (Qwen2-VL) \cite{Med-R1}         & 3B   & OmniMedVQA          & 72.57 \\
GMAI-VL-R1 (Qwen2.5-VL) \cite{GMAI-VL-R1}   & 7B   & GMAI-MMBench (test) & 43.84 \\
\bottomrule
\end{tabular}
\end{table*}

\begin{table*}[htbp]
\centering
\caption{Evaluation Results of Test-time Enhanced Methods (Prompting, RAG, \& Multi-agent). Note: Values represent task-isolated literature reporting across varying settings rather than direct horizontal head-to-head comparisons.}
\label{tab:appendix-test-results}
\setlength{\tabcolsep}{6pt}
\renewcommand{\arraystretch}{1.15}
\begin{tabular}{l c c c}
\toprule
\textbf{Model} & \textbf{Size} & \textbf{Dataset} & \textbf{Metric / Acc.} \\
\midrule
CoD \cite{chen2024cod}                       & 65B  & MMLU                & 63.5 \\
MedAdapter (GPT-4) \cite{shi2024medadapter}        & -    & MMLU                & 87.42 \\
MedAdapter (GPT-3.5) \cite{shi2024medadapter}      & -    & MMLU                & 70.55 \\
MedAdapter (LLaMA-2) \cite{shi2024medadapter}      & 13B  & MMLU                 & 33.96 \\
BioRAG \cite{jeong2024improving}                     & 7B   & MMLU                & 53.9 \\
BioRAG \cite{jeong2024improving}                    & 13B  & MMLU                & 57.2 \\
Clinical CoT (GPT-4) \cite{kwon2023large}       & -    & ADNI                 & 68.4 \\
DR-CoT (Qwen-Instruct) \cite{wu2024large}    & 7B   & HotpotQA            & 72.29 (F1) \\
BP4ER \cite{he2024bp4er}                     & -    & MedDG               & 44.78 (BLEU) \\
MC-CoT \cite{wei2024mc}               & 7B   & PATH-VQA            & 45.07 \\
PathCoT \cite{zhou2025pathcot}                  & -    & PubMed              & 40.9 \\
Self-Consistency (GPT-3.5) \cite{wang2023self} & - & TruthfulQA & 67.7 \\
MedAgents (GPT-3.5) \cite{tang2024medagents}       & -    & MedQA               & 64.1 \\
RARoK (LLaMA2-Chat) \cite{zhan2024rarok}       & 7B   & GenMedGPT-5k        & 70.1 (F1) \\
ArgMed-Agents (GPT-3.5) \cite{hong2024argmed}   & -    & MedQA               & 62.1 \\
DR-CoT \cite{wu2024large}                    & -    & DDXPlus              & 33.5 \\
O1 Replication Journey \cite{huang2025o1replicationjourney} & 72B & MedQA & 86.48 \\
SelfRewardRAG (GPT-3.5) \cite{selfrewardrag2024}   & -    & PubMedQA            & 81.1 \\
ClinicalAgent \cite{yue2024clinicalagentclinicaltrialmultiagent}             & -    & TOP                 & 0.7 \\
\bottomrule
\end{tabular}
\end{table*}

\begin{table*}[htbp]
\centering
\caption{Representative Comparison of Reasoning Performance on MMLU-Pro}
\label{tab:mmlu_pro_comparison}
\resizebox{\textwidth}{!}{
\begin{tabular}{l l c p{5.5cm} l c}
\toprule
\textbf{Model Name} & \textbf{Base Model} & \textbf{Data Size} & \textbf{Data Sources} & \textbf{Training Method} & \textbf{MMLU-Pro} \\
\midrule
MedReason (w/ KG Path) \cite{MedReason} & LLaMA-3.1-8B-Instruct & SFT 32k & MedQA, MedMCQA, MMLU, PubMedQA, MedXpert, HLE & SFT + KG & 63.1 \\
MedReason (w/o KG Path) \cite{MedReason} & LLaMA-3.1-8B-Instruct & SFT 32k & MedQA, MedMCQA, MMLU, PubMedQA, MedXpert, HLE & SFT & 59.9 \\
MedReason (w/ KG Path) \cite{MedReason} & Mistral-Instruct-7B & SFT 32k & MedQA, MedMCQA, MMLU, PubMedQA, MedXpert, HLE & SFT + KG & 50.8 \\
MedReason (w/o KG Path) \cite{MedReason} & Mistral-Instruct-7B & SFT 32k & MedQA, MedMCQA, MMLU, PubMedQA, MedXpert, HLE & SFT & 47.6 \\
HuatuoGPT-o1-8B \cite{HuatuoGPT-o1} & LLaMA-3.1-8B-Instruct & SFT 29k & GPT-4o verifiable problems, MedQA, MedMCQA, MMLU-Pro & SFT & 58.7 \\
HuatuoGPT-o1-70B \cite{HuatuoGPT-o1} & LLaMA-3.1-70B-Instruct & SFT 29k + PPO 20k & GPT-4o verifiable problems, MedQA, MedMCQA, MMLU-Pro & SFT + PPO & \textbf{71.0} \\
\bottomrule
\end{tabular}
}
\end{table*}

\begin{table*}[htbp]
\centering
\caption{Performance Progression via Advanced Training Paradigms on MMLU}
\label{tab:mmlu_training_progression}
\resizebox{\textwidth}{!}{
\begin{tabular}{l l c p{5.5cm} l c}
\toprule
\textbf{Model Name} & \textbf{Base Model} & \textbf{Data Size} & \textbf{Data Sources} & \textbf{Training Method} & \textbf{MMLU} \\
\midrule
\textbf{Med-PaLM 2} \cite{Medpalm2} & PaLM 2 (540B) & SFT 193k & MedMCQA & SFT & \textbf{92.0} \\
\textbf{FineMedLM} \cite{FineMedLM-o1} & LLaMA-3.1-8B-Instruct & SFT 264k & FineFineWeb & SFT & \textbf{71.44} \\
\textbf{FineMedLM-o1} \cite{FineMedLM-o1} & LLaMA-3.1-8B-Instruct & SFT 264k + DPO 33k & FineFineWeb, QwQ-32B positive and self-negative & SFT + DPO & \textbf{78.24} \\
\textbf{FineMedLM-o1 (+TTT)} \cite{FineMedLM-o1} & LLaMA-3.1-8B-Instruct & SFT 264k + DPO 33k + TTT & FineFineWeb, QwQ-32B positive and self-negative & SFT + DPO + TTT & \textbf{81.36} \\
\textbf{BioMed-R1 (8B)} \cite{sft/BioMed-R1} & LLaMA-3.1-8B-Instruct & SFT 23k + GRPO 7,627 + adversarial 25\% & m23k reasoning, DeepSeek-R1 (MedQA, MCQA, HeadQA) & SFT + GRPO + adversarial & \textbf{71.1} \\
\textbf{BioMed-R1 (32B)} \cite{sft/BioMed-R1} & Qwen2.5-32B-Instruct & SFT 23k + GRPO 7,627 + adversarial 25\% & m23k reasoning, DeepSeek-R1 (MedQA, MCQA, HeadQA) & SFT + GRPO + adversarial & \textbf{80.7} \\
\bottomrule
\end{tabular}
}
\end{table*}

\end{document}